\documentclass[runningheads]{llncs}
\usepackage{graphicx}
\usepackage{xcolor}

\providecommand\doi[1]{\href{https://doi.org/#1}{\url{#1}}}

\begin{document}
\title{Object Tracking through Residual and \\Dense LSTMs}

\author{Fabio Garcea\inst{1}\orcidID{0000-0003-3460-5297} \and
Alessandro Cucco\inst{1} \and
Lia Morra\inst{1}\orcidID{0000-0003-2122-7178} \and
Fabrizio Lamberti\inst{1}\orcidID{0000-0001-7703-1372}}
\authorrunning{F. Garcea et al.}
\institute{Politecnico di Torino, Dipartimento di Automatica e Informatica, Turin, Italy
\email{\{fabio.garcea, lia.morra, fabrizio.lamberti\}@polito.it, alessandro.cucco@studenti.polito.it}}
\maketitle              % typeset the header of the contribution
\begin{abstract}
Visual object tracking task is constantly gaining importance in several fields of application as traffic monitoring, robotics, and surveillance, to name a few. Dealing with changes in the appearance of the tracked object is paramount to achieve high tracking accuracy, and is usually achieved by continually learning features. Recently, deep learning-based trackers based on LSTMs (Long Short-Term Memory) recurrent neural networks have emerged as a powerful alternative, bypassing the need to retrain the feature extraction in an online fashion. Inspired by the success of residual and dense networks in image recognition, we propose here to enhance the capabilities of hybrid trackers using residual and/or dense LSTMs. By introducing skip connections, it is possible to increase the depth of the architecture while ensuring a fast convergence. Experimental results on the Re$^{3}$ tracker show that DenseLSTMs outperform Residual and regular LSTM, and offer a higher resilience to nuisances such as occlusions and out-of-view objects. Our case study supports the adoption of residual-based RNNs for enhancing the robustness of other trackers.

\end{abstract}
\section{Introduction}
Visual object tracking plays a fundamental role in many applications including, e.g., robotics and video-surveillance. In this paper, we specifically focus on the problem of \textit{generic} object tracking, which can be concisely phrased as follows: ``given a bounding box enclosing an arbitrary object at time $t$, produce bounding boxes for that object in all future frames'' \cite{gordon2018re}.

Current generic 2D image tracking systems predominantly rely on training a tracker online according to the \textit{tracking-by-detection} paradigm: an object-specific detector is continuously updated with the new object's aspect at every frame, to cope with changes in shape and appearance, as well as with occlusions, while the object moves. Compared with trackers trained completely offline, this approach is more robust and flexible, but these advantages are paid with a decrease in the frame rate that can be achieved. In recent years, hybrid trackers that combine convolutional neural networks (CNNs) for visual feature extraction with Long Short-Term Memory (LSTM) recurrent neural networks have been widely adopted. An example is represented by the Re$^{3}$ tracker \cite{gordon2018re}: the CNN is trained completely offline, thus reducing the computational load at inference time, and the LSTM is trained to update and store an object-specific model. This method has shown increased accuracy and robustness against comparable trackers, especially during occlusions, but is still sensitive to changes in the object's appearance due to occlusions or partially out-of-view targets, to proximity with similar objects, as well as to the presence of background clutter. 

A possible way to improve the performance of LSTM-based trackers is to increase the complexity of the recurrent module, e.g., by stacking several LSTMs. This approach, however, can make it harder for the training procedure to converge, due to the increased network depth. Inspired by the success of Residual Networks \cite{he2015deep} and Dense Networks \cite{huang2016densely} in image recognition, few works in literature have explored the use of residual connections in LSTMs, mostly for speech and text analysis \cite{kim2018residual,kim2017residual,ding2018densely,gao2018densely,wang2018using}. 

In principle, using deeper and more complex LSTM modules should improve the capability of the tracker to model long-term change sequences. Our contribution is thus the design and experimental validation of Dense and Residual LSTM modules for visual object tracking. To assess, \textit{ceteris paribus}, the added benefit of residual connections in object tracking, we modified the established architecture of the Re$^{3}$ tracker \cite{gordon2018re}. Our experimental evaluation on the OTB50 and OTB100 benchmarks shows that Dense LSTM modules achieve higher robustness to occlusion and out-of-view targets while maintaining a similar parameter count compared to solutions adopting plain, non-residual layouts. 

The rest of the paper is organized as follows. In Section \ref{sec:related work}, related work related to object tracking and residual networks is presented. Afterwards, in Section \ref{sec:network}, we examine the tracker selected as the baseline and propose two different variations of the original layout involving residual connections in the recurrent module. In Section \ref{sec:experiments}, we present the performance obtained by the proposed architectures on different benchmarks and compare them with state-of-the-art trackers. Finally, in Section\ref{sec:conclusions}, we discuss the main findings of our experiments and give some directions for future works.

\section{Related work}
\label{sec:related work}

\subsection{Object tracking}
Modern trackers can be roughly divided in \textit{offline-trained}, \textit{online-trained}, and \textit{hybrid} \cite{ali2016visual,gordon2018re}. Online trackers operate online, continually learning features to update the object's appearance during tracking: trackers adopting the well-known tracking-by-detection paradigm belong to this category. This type of tracker must carefully balance adaptation with real-time response abilities. 

Recent works have exploited the capabilities of deep neural networks (DNNs) to learn from massive amounts of data by training CNN-based trackers completely offline \cite{bertinetto2016fully}. These solutions rely on pre-trained CNNs for feature extraction and can operate at faster than real-time speed, but are intrinsically limited in coping with changes in objects' appearance due to movements, occlusions, blurring, etc.

Hybrid solutions like MDNet \cite{nam2015learning} and Re$^{3}$  \cite{gordon2018re} represent an attempt to merge best qualities from both offline- and online-trained solutions. In the Re$^{3}$ architecture, a CNN is trained offline to perform feature extraction, coupled with an LSTM module that keeps track of the object history over time. A multi-resolution approach is used by combining high-level features derived from the full CNN to low-level features learned by the previous layers, thus increasing the robustness of the feature extraction. This architecture represents a good trade-off between fast, real-time tracking (it achieves speeds of 150 frames per second) and robustness against nuisances such as occlusions.

\subsection{Residual networks}

Residual networks and, in more recent times, densely connected networks have shown superior accuracy and training properties than traditional sequential CNNs, and have consistently achieved state-of-the-art results in image classification and other visual tasks \cite{he2015convolutional,he2016identity}. Densely connected CNNs, or DenseNets, represent an extension of the concept of skip connections: the output from each layer is passed as input to all subsequent layers and, as a consequence of the greater flexibility, have proven more effective than ResNets on a variety of visual tasks.
A question that naturally arises is whether residual connections can prove as beneficial also for LSTM networks and, by extension, if the performance of hybrid trackers can be improved as well. 

The idea of stacking LSTMs in a residual fashion has already been adopted in other fields of study such as distant speech recognition \cite{kim2017residual}, sentiment intensity prediction \cite{wang2018using}, and object tracking \cite{kim2018residual}. Dense LSTMs stacking has been recently explored for sentence classification \cite{ding2018densely} and speech enhancement \cite{gao2018densely}. In \cite{kim2018residual}, a rule-based residual RLSTM has been applied to tracking, achieving good results compared to other state-of-the-art trackers. However, given the complexity of object tracking networks and considering the role played by the feature extraction part (based on convolutional layers), by the training algorithm and by the training set, we believe that only by conducting controlled experiments the impact of residual connections can be fully appreciated. To the best of our knowledge, the role of Dense LSTMs in the context of tracking has not been investigated yet.

\section{ResidualRe$^{3}$ and DenseRe$^{3}$}
\label{sec:network}
\begin{figure}
\begin{center}
\includegraphics[width=\textwidth]{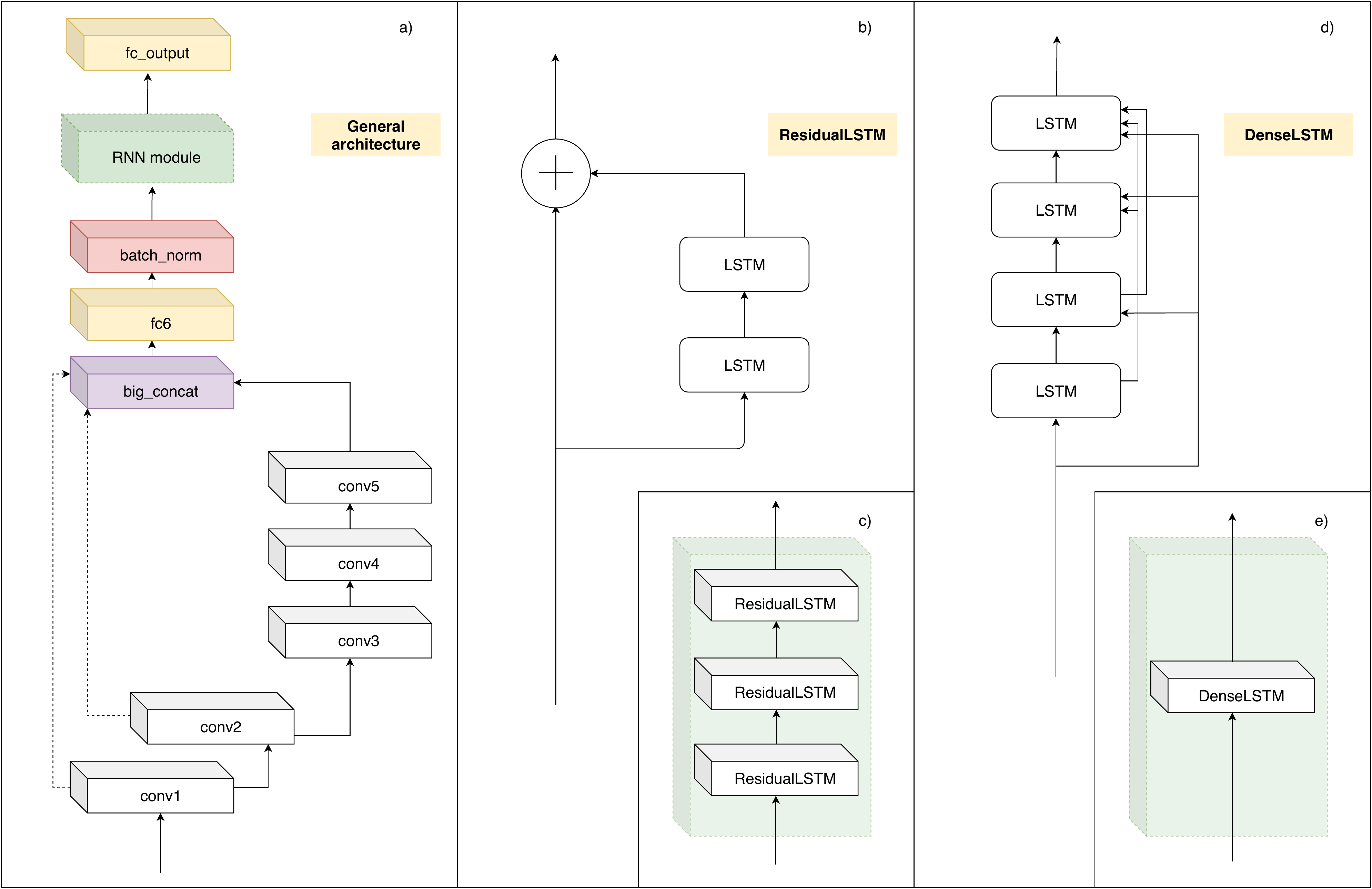}
\caption{A visual comparison of the proposed LSTM-based blocks. a) General structure of the Re$^{3}$ tracker; the main difference between Re$^{3}$, ResidualRe$^{3}$ and DenseRe$^{3}$ is the RNN module (in green). b) Basic structure of the ResidualLSTM tracker, consisting of a series of two LSTMs; in the ResidualRe$^{3}$ alternative shown in c), a stack of three ResidualLSTM blocks has been deployed as the recurrent module. d) Basic structure of the DenseRe$^{3}$ tracker where four LSTMs have been densely connected through skip connections; resulting structure used as recurrent module is reported in e).} \label{fig1}
\end{center}
\end{figure}

For our experiments, we selected the Re$^{3}$ tracker as baseline architecture, and propose two alternative LSTM modules: a ResidualLSTM block consisting of two cascaded LSTMs, and a DenseLSTM block in which four sequential LSTMs are densely connected by applying the same intuition used in DenseNets. The blocks, as well as their position in the overall architecture, are illustrated in Fig. \ref{fig1}. The number and size of layers were carefully chosen to keep the parameter count and the combined depth as similar as possible to the original Re$^{3}$ architecture. 
In the following sub-sections, the main characteristics of the two solutions will be illustrated.

\subsection{Re$^{3}$}
The Re$^{3}$ tracker was firstly proposed in \cite{gordon2018re}. It represents a hybrid solution to the problem of generic object tracking. The layout of this network can be mainly split into three modules solving different tasks. The first module is a stack of convolutional layers used to extract the embeddings from the object being tracked; a concatenation layer is fed with both low-level and high-level information to obtain a more complete representation of the object. In the second module, a recurrent block consisting of a stack of two LSTM layers, each one receiving the features extracted at the previous stage, can keep track of subsequent object's positions and transformations. Finally, a regression layer is used to predict the bounding box of the object in the current frame. The full model is fed with two frame crops from the sequence at each time step; one of them is centered at the object's position in the previous frame, whereas the other is centered at the same position but in the current frame. Both the crops are still large enough to carry some information about the background. 

\subsection{ResidualLSTM-based RNN}
In the ResidualRe$^{3}$ version of the tracker, a different architecture has been adopted for the recurrent module. In particular, a sequence of two LSTMs connected in series has been added to the input of the first LSTM module in a residual block fashion. We will refer to this structure as the ResidualLSTM block. The full layout of the recurrent module consists of a stack of three ResidualLSTM blocks. Since the outputs from both the convolutional module and the LSTMs are summed through a merge layer, they need to share the same number of units. In the original version of the tracker, the CNN output is set to use 1024 units, but we decided to downscale it to 768 units to keep the parameter count of the complete network comparable to that in the original version of the tracker. Moreover, a batch normalization layer has been added after the fully connected layer of the CNN to make its output comparable to that of the first ResidualLSTM block.

\subsection{DenseLSTM-based RNN}
In the DenseLSTM version of the tracker, we decided to replace the recurrent module with a different structure exploiting dense residual connections; a stack of four LSTMs has been densely connected through skip connections, thus allowing each subsequent module to be fed from the output of the previous ones. We will refer to this structure as the DenseLSTM block. In this case, the fully connected layer on top of the CNN has been set to use 900 units instead of the original 1024 units to keep a low parameter count for the following recurrent module; moreover, a batch normalization layer has been added on top of this layer to speed up model convergence. In the full network layout, we used a single DenseLSTM block composed of four LSTMs with 512 units each. With these constraints, we were able to maintain a complexity similar to that of the original Re$^{3}$ tracker.

\subsection{Training and implementation details}
The training procedure is the same as in the original Re$^{3}$ paper \cite{gordon2018re}. We here summarize the most important steps. Before starting the training, synthetic data are produced with several augmentation techniques such as horizontal flipping and random noise generation, and weights from AlexNet are loaded in the CNN; LSTM states are initialized to zero, whereas other weights are set using MSRA initialization. The adopted optimizer is Adam with momentum and weight decay set to default values, and a learning rate decreasing from $10^{-5}$ to $10^{-6}$ after 10.000 iterations. Finally, the loss function is the Mean Absolute Error (MAE), and the number of iterations is 200.000. The training is initialized with 64 batch size, 2 unrolls and 1 probability of using the ground truth bounding boxes as a reference to crop the frame at the following time step; as soon as the loss plateaus, the batch size is halved and the unrolls are doubled (up to 32 unrolls). Moreover, the probability (initially 0) of mixing the predicted bounding boxes with the ground truth is increased using steps of 0.25; in this way, the network can learn from its errors during training thus being able to partly recover from errors at test time.

Since reproducibility of deep learning models is notoriously difficult to achieve, being the training procedure inherently random and affected by several factors including the training environment  \cite{bouthillier2019unreproducible,marrone2019reproducibility}, the original model has been retrained following the steps in the original publication on the ILSVRC2014 DET and ILSVRC2017 VID datasets, starting from the original code provided by the authors. 

The training procedure for the modified networks followed the same used for training the Re$^{3}$ original version with some minor changes. For the ResidualRe$^{3}$ tracker, a faster learning rate of $10^{-4}$ was initially set, then reduced to $10^{-5}$ and $10^{-6}$ when noticing that the loss function starts to plateau; moreover, a faster learning rate scaling of $10^{-1}$ was adopted for the finetuning of the CNN module weights. For the DenseRe$^{3}$ network, the procedure was similar, but we started to increase the probability of using the network prediction only after 32 unrolls and each time the loss function showed a plateau. 

All the experiments were performed on a system configured with an i7 2600 CPU, 8GB DDR3 1333 MHz RAM and an NVIDIA GTX 1060 3GB GPU.

\section{Experimental results}
\label{sec:experiments}
First, we report the results of training and testing all the architectures on the ILSVRC2014 DET and ILSVRC2017 VID datasets, considering also the re-trained Re$^{3}$ model provided by \cite{gordon2018re}. Secondly, we compare results obtained by the original published Re$^{3}$ model on the challenging OTB50 and OTB100 benchmarks \cite{WuLimYang13} along with several state-of-the-art trackers. The OTB100 benchmark consists of 100 different image sequences reporting objects from different classes and assignable to different attributes (occlusion, motion blur, out-of-view, etc.).

\subsection{Training  ResidualRe$^{3}$ and  DenseRe$^{3}$}
The results for the architectures under test are reported in terms of two different metrics, namely, the number of targets lost by the tracker, and the Mean Intersection Over Union (IOU) between the predicted and the ground truth bounding boxes. The parameters count is reported as well, to highlight how the new architectures remain comparable to the model from the original paper.

Our results show an improvement in the Mean IOU score and a lower number of lost targets compared to the retrained version of Re$^{3}$, while keeping a comparable parameter count (Table  \ref{tab1}). The evaluation of training results for DenseRe$^{3}$ showed significant improvements compared to the Retrained version of Re$^{3}$ and ResidualRe$^{3}$ (Table \ref{tab1}), thus demonstrating the advantages brought by the DenseLSTM blocks.

\begin{table}
\begin{center}
\caption{Training results for different Re$^{3}$ architectures. It should be noticed how residual and dense LSTM improve performance with minimal increase in parameter count.}\label{tab1}
\begin{tabular}{|cccc|}
\hline
\textbf{Tracker}     & \textbf{Lost targets} & \textbf{Mean IOU} & \textbf{Parameter count} \\ \hline
Re$^{3}$ (retrained) & 350                   & 0.64              & 85.699.686               \\ \hline
ResidualRe$^{3}$     & 303                   & 0.66              & 87.716.712               \\ \hline
DenseRe$^{3}$        & 258                   & 0.68              & 87.031.408               \\ \hline
\end{tabular}
\end{center}
\end{table}

Concerning the original Re$^{3}$ architecture, it is worth noticing that we did not achieve the same performance of the model released by the authors even though, to the best of our knowledge, we followed the same training curriculum for Re$^{3}$. Specifically, our retrained model achieves 350 lost targets with a Mean IOU of 0.64 versus the 243 lost targets and 0.72 Mean IOU for the weights provided by the authors. 

This discrepancy is not entirely surprising, since reproducing results is a well-known issue of deep learning-related research, and maybe due to slight differences in implementation or training parameters. Recent research also highlighted the effect of random initialization on the estimated performance of image classification networks \cite{bouthillier2019unreproducible}; however, for complex deep learning architectures, such as object trackers, running multiple experiments per configuration requires substantial computational resources. In the future, we plan on exploring this issue in more detail. For the remaining experiments, we compare DenseRe$^{3}$ with the original model provided by the authors, which albeit less favourable allows an easier comparison with the previous literature.

\subsection{Benchmarks evaluation}
Since we needed a state-of-the-art reference tracker to compare our results, we opted for the Recurrent Filter Learning tracker (RFL) \cite{yang2017recurrent}; besides its high performances, this model is based on a recurrent module thus representing an appropriate reference architecture for our experiments.

\begin{figure}[!ht]
\begin{center}
\includegraphics[width=\textwidth]{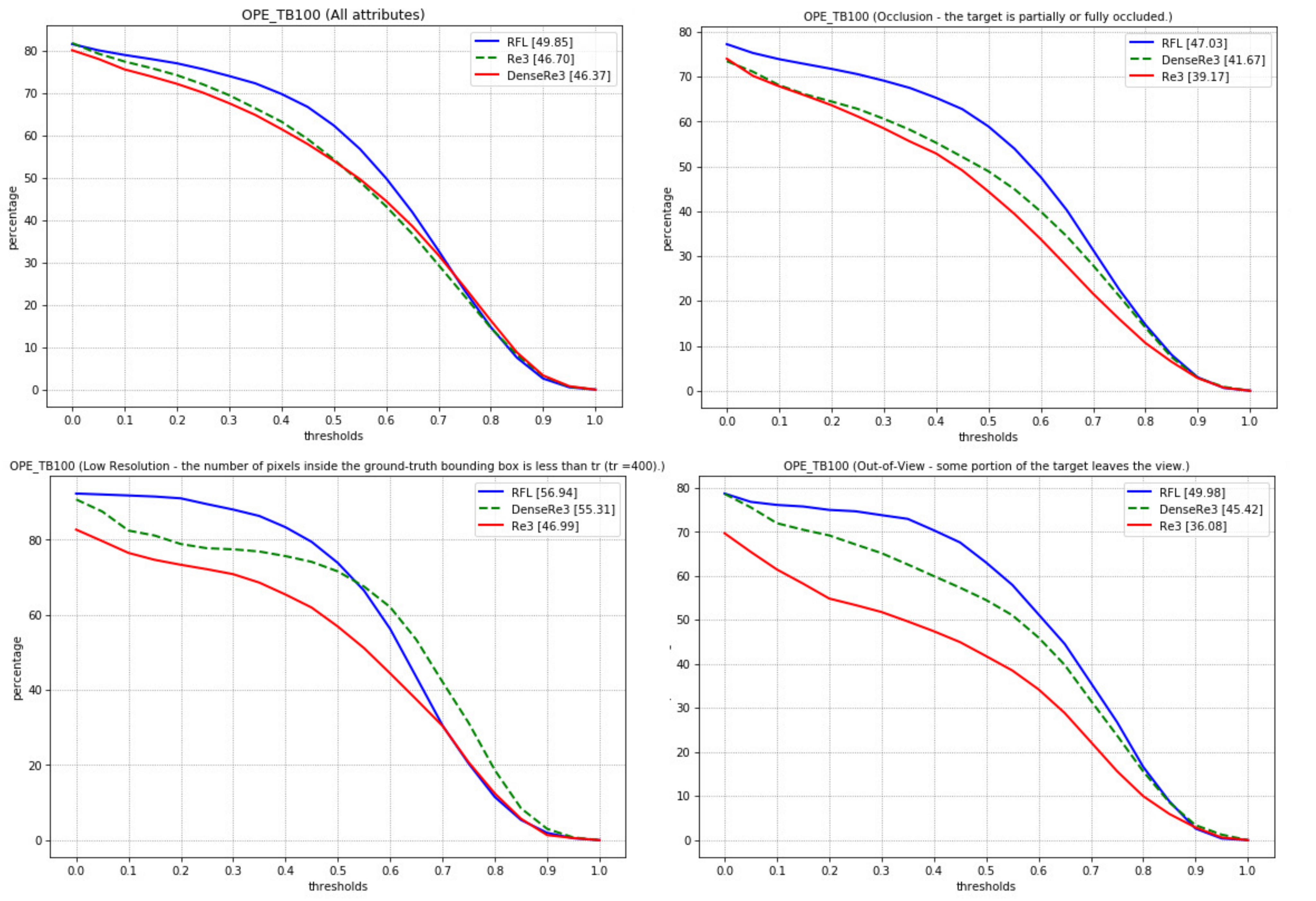}
\caption{Results on the OPE TB100 benchmark for RFL, the original version of Re$^{3}$, and the proposed DenseRe$^{3}$ architecture. The percentage of frames where the mean IoU is greater than a threshold (y axis) is plotted as a function of the threshold value (x axis). The success plots report the results for different sequence attributes (all attributes, occlusion, low resolution and out-of-view objects). Whilst in some cases the DenseRe$^{3}$ model scores are similar to the z version, in other cases they show a better performance of our architecture.} \label{fig2}
\end{center}
\end{figure}

We report here the results of the One Pass Evaluation (OPE) protocol on the OTB50 and OTB100 benchmarks; while the results on OTB100 benchmark were computed at test time starting from the code as provided by the authors, those for OTB50 were already provided for multiple trackers and thus have not been re-computed. Once the baseline was defined, we executed the OPE TB100 benchmark for the RFL tracker, the original version of Re$^{3}$ and the DenseRe$^{3}$ networks; the results (Fig.~\ref{fig2}) show that our model can outperform the original version of Re$^{3}$ in sequences characterized by low resolution, occlusions and out-of-view objects, while still performing similarly to the original architecture in other cases.

\begin{figure}[!ht]
\begin{center}
\includegraphics[width=\textwidth]{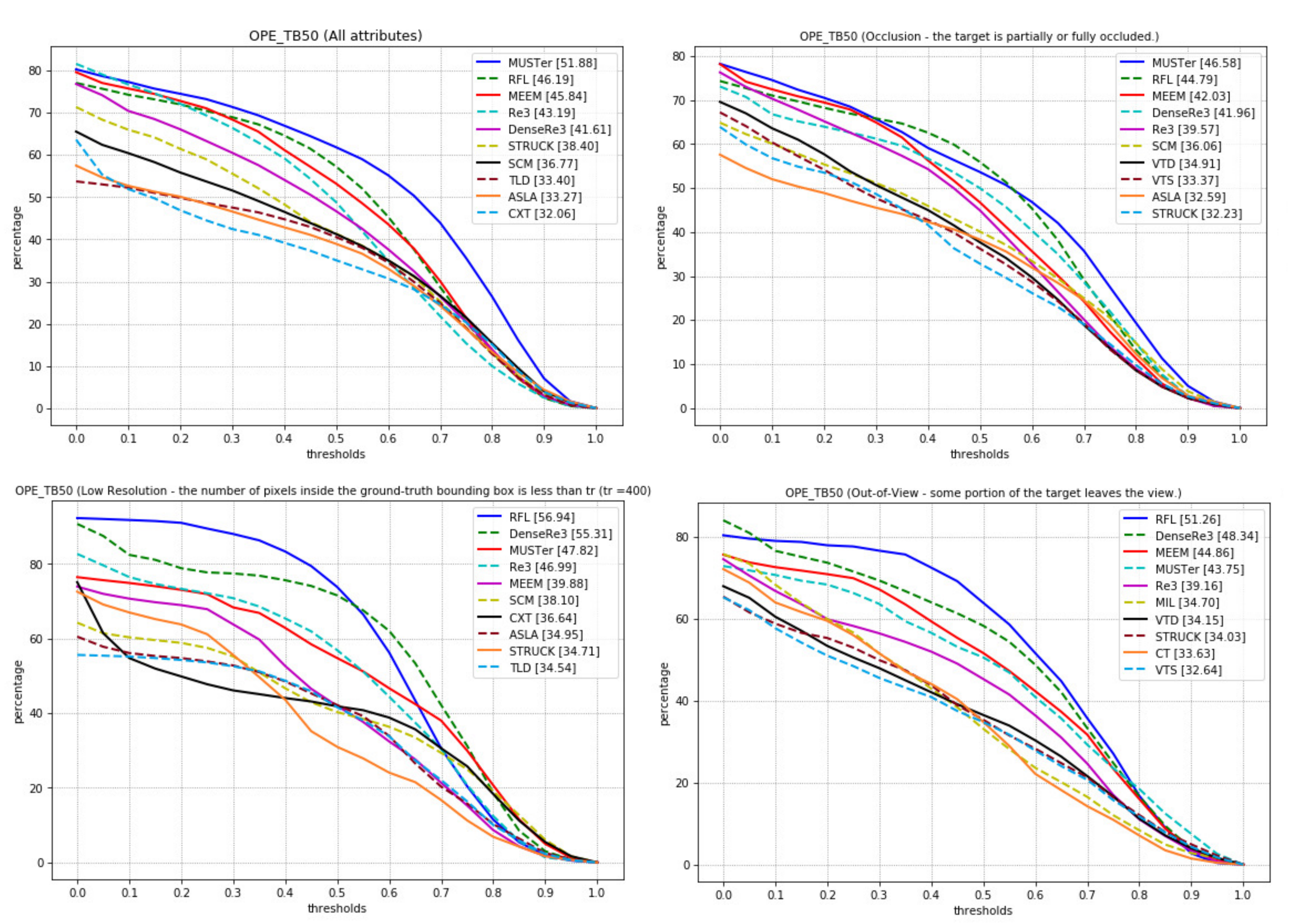}
\caption{Results on the OPE TB50 benchmark for the proposed DenseRe$^{3}$ architecture and other state-of-the-art trackers. The percentage of frames where the mean IoU is greater than a threshold (y axis) is plotted as a function of the threshold value (x axis). The success plots reports the results for different sequence attributes (all attributes, occlusion, low resolution and out-of-view objects). In most of the subsets DenseRe$^{3}$ scores are similar to those obtained by the original Re$^{3}$ tracker. DenseRe$^{3}$ achieves high performance in sequences with low resolution, occlusion and out-of-view target.} \label{fig3}
\end{center}
\end{figure}

We then ran the OTB50 benchmark for the RFL, the original Re$^{3}$, the DenseRe$^{3}$ and other state-of-the-art trackers to evaluate the performances of our model with a larger pool of different architectures. The results (Fig. \ref{fig3}) are aligned with those obtained with OTB100; like with the other benchmark, the model performs particularly well with sequences characterized by the above attributes, reaching the top-4 positions for all attributes. In all the other sequences, it performs worse than Re$^{3}$, even though it can reach the top-5 positions.

\begin{figure}[!ht]
\begin{center}
\includegraphics[width=\textwidth]{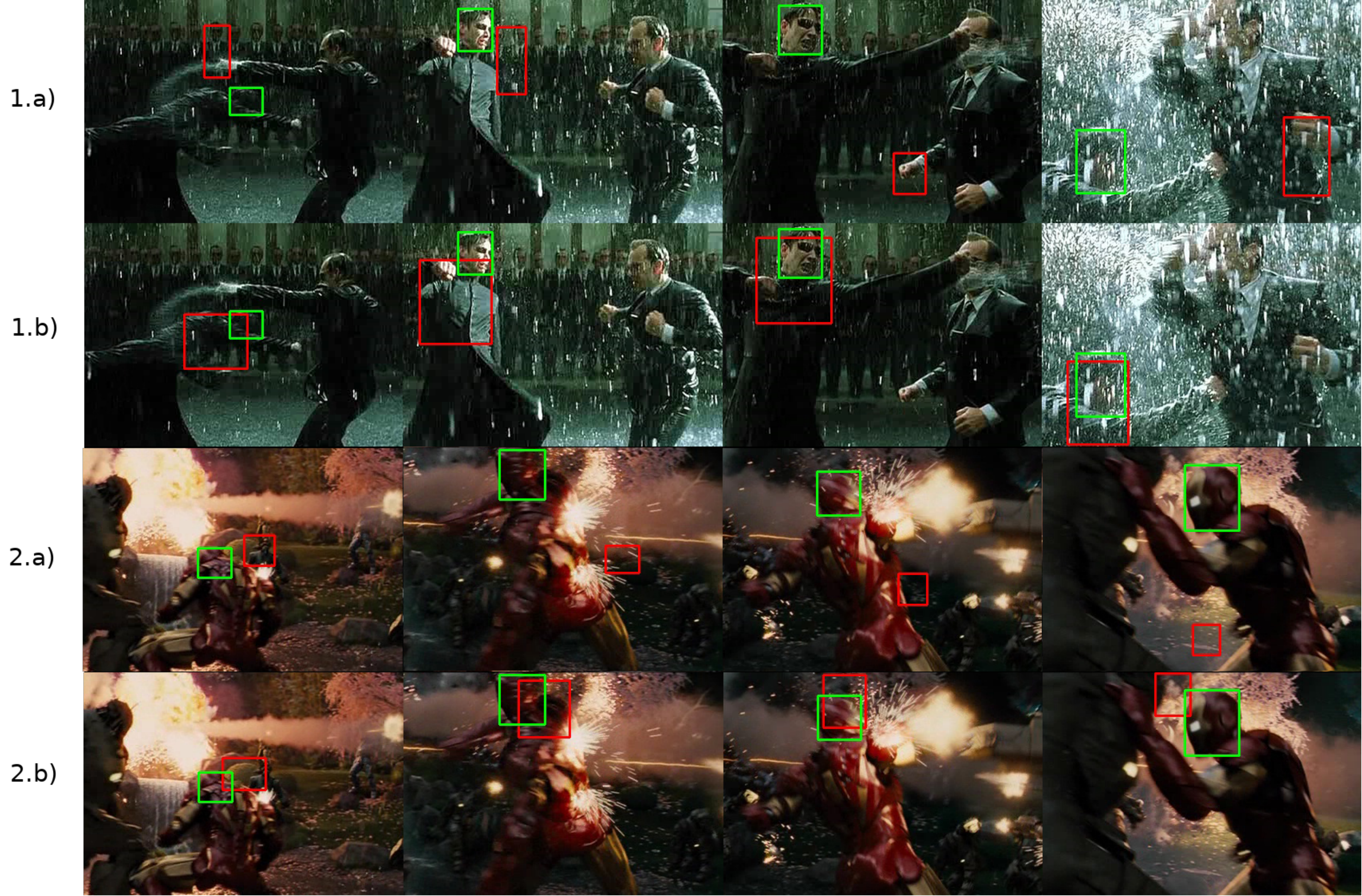}
\caption{Example sequences from the OPE TB100 benchmark evaluated on DenseRe$^{3}$ and plain Re$^{3}$. The first sequence, named "Matrix", has been annotated with both Re$^{3}$ (1.a) and DenseRe$^{3}$ (1.b). Similarly the sequence named "Ironman" has been annotated by both Re$^{3}$ (2.a) and DenseRe$^{3}$ (2.b). The bounding boxes annotated by DenseRe$^{3}$ intersect with the ground-truth (green bounding box) also in case of disturbances showing thus a robust behavior of the network if compared to plain Re$^{3}$.} \label{fig5}
\end{center}
\end{figure}

Finally, two example sequences from the OTB100 benchmark are reported (see Figure \ref{fig5}) to depict, in a visual fashion, the achieved improvement. The first sequence, named "Matrix", is characterized by multiple attributes like occlusion, fast motion and illumination variation. The second sequence, named "Ironman", similarly presents multiple attributes as well, such as occlusion and out-of-view. Both the sequences have been annotated by Re$^{3}$ and DenseRe$^{3}$ with a red bounding-box representing the prediction of the tracker under test for each frame. 

It's evident how, in the sequence "Matrix", Re$^{3}$ (sub-sequence 1.a) loses the track of the object due to the fast motion of the body and the partial occlusion of the face features and consequently starts to track the hand of the second character. On the other side, DenseRe$^{3}$ (sub-sequence 1.b) can keep track of the object also in presence of disturbances and it's able to progressively recover from the error caused by the occluded frames. Moreover, a robust behavior can be appreciated in the fourth frame of the sequence where the model keeps track of the object in the presence of an important variation in the illumination.

In the second example, Re$^{3}$ (sub-sequence 2.a) loses the track of the object when it goes temporary out-of-the-view and it's not able to recover from its own errors thus ending up losing track of the object. On the other hand, DenseRe$^{3}$ (sub-sequence 2.b) shows again a robust behavior in case of occluded and out-of-view frames being able to keep track of the object even if with some difficulties due to the complexity of the frame.

\section{Conclusions and future work}
\label{sec:conclusions}
In this work, we explored the potential benefit of Residual and Dense LSTM in hybrid object tracking architectures. The idea of introducing residual and dense skip connections in LSTMs has been successfully explored in other applications, such as speech and text recognition. 

We here investigate a case study in object tracking, in which we modified the architecture of an LSTM-based tracker, the  Re$^{3}$ architecture, using both ResidualLSTM and DenseLSTM modules. Our experiments showed that both ResidualLSTM and DenseLSTM modules can be successfully used to enhance the robustness of the Re$^{3}$ tracker, as in low resolution or occlusion attributes, while keeping a parameter count comparable to the original version. 
In general the proposed architecture appears to be more robust to the presence of occlusions, low resolution and other disturbances. Residual and even more dense architecture allow to connect each layer not only with the previous layer, but also with previous ones. Skip connections are an essential component of deep convolutional neural networks allowing to increase the number of layers without incurring in vanishing gradients or other numerical instability. DenseRe$^{3}$ is characterized by four LSTM blocks, instead of the two blocks of the plain  Re$^{3}$ tracker, thus effectively doubling the depth of the network. Nonetheless, the use of skip connections makes the information flow across the layers easier ensuring fast convergence, thus increasing performance in a comparable number of iterations. Previous works reported that increasing the number of layers in plain LSTM may lead to performance degradation, however,  this phenomenon can be reduced or even reversed when residual connections are introduced \cite{wang2018using}. We observed an even greater benefit from dense connections, but the relationship between performance and depth should be analyzed in a more systematic fashion. 

We also hypothesize that in a densely connected structure, where the activations of each layer are fed to all subsequent ones, it is possible to more effectively "remember" the history of the object being tracked, thus improving the robustness in the presence of occlusions and background clutter, or when the object moves out-of-view. 

We expect that similar improvements could be found on other architectures currently relying on plain LSTM modules. In the future, we plan to explore the advantages of ResidualLSTM and DenseLSTM blocks in other trackers or other visual tasks.

\end{document}